\documentclass[review]{elsarticle}

\usepackage{lineno,hyperref}
\modulolinenumbers[5]
\usepackage{adjustbox}
\usepackage{amsmath,bm}
\usepackage{amssymb}
\usepackage{siunitx}
\usepackage{textcomp}
\usepackage{mathtools}
\usepackage{tabularx,multirow,makecell}
\usepackage[ruled,vlined]{algorithm2e}
\usepackage{setspace}
\usepackage{xcolor}
\usepackage{color}\usepackage{color}

\usepackage{soul}
\usepackage{booktabs}
\usepackage{makecell}

\newcommand{\cmmnt}[1]{}

\usepackage[flushleft]{threeparttable}

\usepackage{setspace}
\usepackage{xcolor}
\usepackage{color}
\usepackage{soul}

\usepackage{leftidx}

\def\XS{\xspace}
\DeclareMathAlphabet{\mathb}{OML}{cmm}{b}{it}
\def\sbm#1{\ensuremath{\mathb{#1}}}                
\def\scu#1{\ensuremath{\mathcal{#1\XS}}}           

\def\figureabvr{Figure\XS} 
\def\tableabvr{Table\XS}
\def\eg{\textit{e.g.,}\XS}
\def\etal{\textit{et al.}\XS}
\def\etc{\textit{etc.}\xspace}

\def\Ib{{\sbm{I}}\XS}  
\def\Pb{{\sbm{P}}\XS}  \def\pb{{\sbm{p}}\XS}
\def\Gc{{\scu{G}}\XS}   
\def\Lc{{\scu{L}}\XS}   
\def\Vc{{\scu{V}}\XS}

  \def\fb{{\sbm{f}}\XS}
  \def\gb{{\sbm{g}}\XS}
   
  \def\mb{{\sbm{m}}\XS}
\def\Wb{{\sbm{W}}\XS}  
  \def\bb{{\sbm{b}}\XS}

\newcommand*{\Scale}[2][4]{\scalebox{#1}{$#2$}}%

\journal{}

\begin{document}
\begin{frontmatter}

\title{Graph Self-Supervised Learning for Endoscopic Image Matching}

\author[crns,smart]{Manel Farhat}
\ead{farhat.manel@hotmail.fr}
\author[crns,smart]{Achraf Ben-Hamadou\corref{mycorrespondingauthor}}
\ead{achraf.benhamadou@crns.rnrt.tn}
\address[crns]{Centre de Recherche en Numérique de Sfax, Technopôle de Sfax, 3021 Sfax, Tunisia}
\address[smart]{Laboratory of Signals, systems, artificial Intelligence and networks, Technopôle de Sfax, 3021 Sfax, Tunisia}
\cortext[mycorrespondingauthor]{Corresponding author}

\begin{abstract}
Accurate feature matching and correspondence in endoscopic images play a crucial role in various clinical applications, including patient follow-up and rapid anomaly localization through panoramic image generation. However, developing robust and accurate feature matching techniques faces challenges due to the lack of discriminative texture and significant variability between patients. To address these limitations, we propose a novel self-supervised approach that combines Convolutional Neural Networks for capturing local visual appearance and attention-based Graph Neural Networks for modeling spatial relationships between key-points. Our approach is trained in a fully self-supervised scheme without the need for labeled data. Our approach outperforms state-of-the-art handcrafted and deep learning-based methods, demonstrating exceptional performance in terms of precision rate (1) and matching score (99.3\%). We also provide code and materials related to this work, which can be accessed at \href{https://github.com/abenhamadou/graph-self-supervised-learning-for-endoscopic-image-matching}{https://github.com/abenhamadou/graph-self-supervised-learning-for-endoscopic-image-matching}.
\end{abstract}


\end{frontmatter}

\doublespacing
\section{Introduction}
Endoscopy plays an important role in the early detection of diseases that reduce the risk of developing numerous cancers, such as gastric, colorectal, and bladder cancers. It is a nonsurgical clinical procedure consisting of inserting into the person's organs a flexible or rigid tubular instrument called an endoscope equipped with a camera and a light source. This procedure allows doctors to examine and explore the cavity and the inner walls of the organs. However, the reduced field of view of the endoscopes is still a major limitation during endoscopic examination, as it only covers a small surface, and the doctor repeatedly moves the endoscope back and forth to examine the lesions, which are typically spread over larger areas of interest. This makes the assessment of their spatial distribution inaccurate and time-consuming. As a solution to this limitation, a panoramic image can be created from an endoscopic video sequence, which requires detecting overlapping image parts and estimating the image transformation between consecutive image frames.

In the literature, three main families of approaches were investigated to address this task: motion-based methods, global image matching methods, and local image matching methods. Motion-based methods use optical flow-based approaches to find correspondences between endoscopic images \citep{\cmmnt{Hernandez10,6738266}ALI2016425,CHU2020105370, zenteno}. These methods have demonstrated their efficiency in registering consecutive endoscopic image frames. However, their main limitation is their sensitivity to challenging situations such as large viewpoint changes, weak textures, and illumination changes. Global image matching methods use the entire image content (\eg contours, color, graphs, \etc) and ensure the global coherence of panoramic images by penalizing discontinuities and optimizing an explicit smoothness assumption. However, they are memory-intensive, time-consuming, and sensitive to the initial geometric transformation between images. Moreover, the strong hypothesis about organ surface flatness is often not verified. For instance, Miranda-Luna \etal generated a panoramic bladder endoscopic image based on the maximization of mutual information \citep{4360071}, while Weibel \etal used graph-cut techniques to minimize a global energy function computed across the entire endoscopic image pixels \citep{WEIBEL20124138}.

Global matching methods are known to be memory-intensive and time-consuming as they calculate the energy function over all pixels in the image. Moreover, they are better suited for consecutive image matching with minor changes because they are sensitive to the initial geometric transformation between images. Additionally, the strong hypothesis about organ surface flatness is often not verified.

On the other hand, local image matching methods focus on extracting feature vectors from local image regions to find image correspondences. These methods only use local information and do not consider neighboring images regions. They typically start by detecting image key-points, computing discriminative features for each key-point, and finally performing key-point matching by search for nearest-neighbors in the feature space. They are computationally efficient and are, therefore, suitable for real-time applications. SIFT \citep{Lowe2004} and ORB \citep{Rubleeetal2011} are two of the most popular hand-crafted local features, widely adopted for endoscopic image matching tasks \citep{ZHANG2022103261,fnbot.2022.840594,Liu:22}. \cmmnt{9478624}

Recently, deep learning-based methods for image matching have focused on learning discriminative local features using Convolutional Neural Networks (CNNs) \citep{8100132,sosnet2019cvpr,\cmmnt{8578167}Luo2018GeoDescLL,hynet2020}. However, while recent research proves the efficiency of CNNs in local feature extraction, they do not take the larger global context into consideration. Besides the visual appearance of key-points, leveraging more contextual information can improve their distinctiveness. Using the relationship that may exist between key-points in neighborhoods can enhance their representations. The local image matching approach can be enhanced by exploring both the visual appearance of the key-points and their spatial relationships.

Nowadays, Graph Neural Networks (GNNs) are designed for graph-structured data and have demonstrated prominent results. Recently, several advanced GNN entities have been proposed such as the graph convolutional network (GCN) \citep{kipf2017semi} and the graph attention network (GAT) \citep{velivckovic2018graph}. Most of GNN models used the neural message passing technique \citep{gilmer2017neural}, to pass and aggregate node information across the graph.

Graph network representations have the potential to enhance the interpretation of feature representation by modeling the spatial relationships between different key-points. In this regard, some methods have proposed local feature matching as a graph matching problem  \citep{sarlin2020superglue, 4408838, 10.1007/978-3-540-88688-4}. Graph matching methods are commonly used for local feature matching in both 2D \citep{ 4408838, 10.1007/978-3-540-88688-4} and 3D \citep{1544893,bai2021pointdsc} applications. The main objective of graph matching methods is to generate node correspondences across two graphs.

Existing graph neural networks are typically trained using a supervised scheme \citep{10.5555/3367471.3367673,pmlr-v97-wu19e}, which requires a large amount of labeled data for training. The lack of annotated training data is a significant challenge in using graph neural networks for endoscopic image matching. Recently, self-supervised learning of Graph Neural Networks has recently gained attention as a potential solution to this problem. Unlike self-supervised learning in computer vision, which relies on visual data augmentation, self-supervised learning for graphs uses graph augmentation or corruption techniques such as node and edge masking \citep{Kipf2016VariationalGA, 10.1145/3534678.3539321}. These methods are better suited to graph structures. Overall, it is clear that the design of a self-supervised framework that effectively integrates visual appearance and graph structure remains an ongoing challenge.


The main contribution of this paper is a novel self-supervised approach for endoscopic image key-point matching using deep learning techniques to generate panoramic images. Our method is motivated by the idea that the visual appearance of key-points can be captured by a CNN, while the spatial relationships can be learned by a GNN. To this end, we designed a local feature descriptor model that combines CNN and GNN to transform patches extracted around key-points into a discriminative embedding space for effective matching.

The remaining of this paper is divided into four sections. Section \ref{sec:relatedwork} provides an overview of the related work for key-point description based on both handcrafted and deep learning techniques. Then, we present a detailed description of our proposed approach in Section \ref{sec:proposed}. The experimental results are discussed in Section \ref{sec:experiments}, followed by a summary of our findings and suggestions for future work in Section \ref{sec:conclusion}.

\section{Related works}
\label{sec:relatedwork}
The related work can be divided into three main sections. Firstly, we provide an overview of local feature descriptor methods, including both handcrafted features and deep learning-based local descriptors. Secondly, we discuss the use of GNNs for feature matching, focusing on their ability to capture the spatial relationships between features. Finally, we summarize recent developments in graph self-supervised learning approaches.

\subsection{Local feature descriptors}
Handcrafted local feature descriptors were classified into floating and binary descriptors according to the type of their discriminative vector \citep{Metal2021}. The well-known floating point descriptor SIFT \citep{Lowe2004} was the first widely successful attempt at local feature description. Many subsequent researchers focused on improving performances and computational requirements such as the speeded-up robust feature SURF \citep{Bayaetal2008}. In \citep{Alcantarillaetal2012}, authors proposed the KAZE descriptor. They used a detection pipeline similar to the SURF method. However, They were based on the MU–SURF method \citep{Agrawaletal2008} to generate a non-linear scale-space after using a non-linear diffusion filter.

Many binary descriptors, on the other hand, have been proposed to improve the speed of the floating-point descriptors for real-time applications. Here are some examples of these methods: AKAZE \citep{Alcantarillaetal2013}, an accelerated version of the KAZE algorithm, and ORB \citep{Rubleeetal2011}, an improved version of BRIEF \citep{Calonderetal2010} with rotation invariance.

Recently, deep learning approaches have shown significant improvements over handcrafted methods in various computer vision tasks \citep{\cmmnt{8918994} Kim21,Liu20_classification\cmmnt{8451300}}. In the context of image matching, MatchNet \cite{Han2015MatchNetUF} is one of the early successful learning-based methods, which employs a Siamese network consisting of two networks: a feature network for extracting feature representation, and a metric network (comprised of three fully connected layers) for measuring the similarity of feature pairs. To further enhance descriptor performance, DeepDesc \cite{7410379} utilized a Siamese network with L2 distance training, which exclusively selected hard pairs. The L2-Net \cite{8100132}, a network composed of seven convolutional layers, outperformed handcrafted methods, including SIFT. Another method, GeoDesc \cite{Luo2018GeoDescLL}, incorporated geometric constraints from multi-view reconstructions to learn more robust descriptors. Alternatively, the SOSNet model \cite{sosnet2019cvpr} introduced second-order similarity regularization, which optimized the intra- and inter-class distances to improve local descriptor performance. The same authors also proposed HyNet \citep{hynet2020}, a network architecture that is a advanced version of L2-Net, with all Batch Normalisation (BN) layers replaced by Filter Response Normalization (FRN) layers. CNDesc \citep{CNDesc9761761} introduced cross-normalization technology as an alternative to common L2 normalization.

Although deep learning methods based on CNN have shown impressive results in local feature extraction and image matching, the scarcity of labeled training data in the medical domain, particularly for endoscopic image matching, remains a significant challenge for developing such models. In our previous work, we proposed a self-supervised approach for matching endoscopic image key-points, which demonstrated competitive performance compared to state-of-the-art supervised methods \citep{FARHAT2023118696}. However, both local descriptors and our self-supervised approach overlook the spatial-topological relationships among the visual elements of the image scene. To address this limitation, Graph Neural Networks (GNNs) have emerged as a popular technique for learning topological and spatial relationships in graphs \cite{9046288,\cmmnt{rs13071404,}rs13010119}.


\subsection{Graph Neural Network for feature matching}
Considering the fact that humans find correspondences between images not only based on the local appearance of the key-point but also by considering the large global context, some methods introduce other contextual information, such as the spatial relationship between visual elements, to improve the image matching process \cite{\cmmnt{rs12234003},rs14061478,9892682}. As a recent advancement in artificial intelligence, the graph neural network (GNN), first introduced in \cite{1555942}, is intended to handle graph-structured data and has demonstrated prominent performance. Many computer vision tasks can be described as graph matching problems such as image classification \cite{9010734,ML-GCN_CVPR_2019}, object detection \cite{9561110, BaldassarreSSA20},  and image matching \cite{sarlin2020superglue,sun2021loftr}. For the image matching task, the graph nodes correspond to the local features of the image, and the graph edges correspond to the relationship aspects between features. Graph nodes and edges can be attributed to encode the feature vectors. Thus, image matching consists of a graph matching process and finding a correspondence between nodes of the two graphs.

In this context, Wiles \etal \cite{Wiles2021CVPR} introduces a new image matching approach based on a spatial attention mechanism (co-attention module (CoAM)). In current research \cite{sarlin2020superglue}, authors proposed SuperGlue, a new neural architecture for image matching. Inspired by the transformer architecture \cite{10.5555/3295222.3295349}, SuperGlue used attention-based graph neural networks, it used self and cross-attention to take advantage of both the visual appearance and the spatial relationships of the key-points. Similar but different to SuperGlue, LoFTR (Local Feature TRansformer) \citep{sun2021loftr} used self and cross-attention layers based on a linear transformer to optimize computational complexity.

\subsection{Graph self-supervised learning}

Motivated by the success of self-supervised learning in natural language processing \citep{devlin-etal-2019-bert,Ryu2021REAA} and in computer vision \citep{misra2020self,jing2020self\cmmnt{goyal2019scaling}}, the research community has become increasingly interested in self-supervised learning of graph neural networks. According to \citep{9764632}, graph self-supervised learning methods can be classified into two categories: predictive or generative methods and contrastive-based methods.

The predictive methods aim to reconstruct and predict missing parts of the input graph \citep{kim2021find,\cmmnt{park2019symmetric,}hasanzadeh2019semi}. 
The prediction targets are some elements of the provided graphs such as some edges or some attribute nodes. In \citep{Kipf2016VariationalGA}, authors proposed a self-supervised graph autoencoder called VGAE, it reconstructs the adjacency matrix by predicting missing edges. GraphMAE \citep{10.1145/3534678.3539321} focused on reconstructing masked attributes of some graph nodes. GPT-GNN \citep{gpt_gnn} used both edge and node reconstruction iteratively based on an autoregressive framework.

Contrastive-based methods are mainly founded on the mutual information maximization concept \citep{hjelm2019learning}. The key idea behind the contrastive-based methods for graph self-supervised learning is to maximize the mutual information between similar graph instances (positive samples) while minimizing the mutual information between graph instances with unrelated information (negative samples). DGI  \citep{velickovic2018deep} extended DIM \citep{10.5555/3454287.3455679} to graphs. It proposed the first contrastive method based on the maximization of the mutual information to graph-structured inputs to obtain a more discriminative node embedding.  

Similar to but different from DGI which evaluated only the node-level embeddings, InfoGraph \citep{Sun2020InfoGraph} aimed to create an expressive representation at the whole graph level. In \citep{10.5555/3524938.3525323}, authors proposed a self-supervised method that sought to learn both node and graph representations by maximizing the mutual information between multi-views of graphs and they demonstrated that, unlike visual representation, using more than two graph views does not increase the performance. GRACE \cite{Zhu:2020vf}  introduced a contrastive learning approach to minimize the agreement of node embeddings between two graphs views generated by node corruption. In GraphCL \cite{NEURIPS2020_3fe23034}, the authors designed four types of graph augmentations for learning graph-level representation and showed that the proposed approach could improve the performance of downstream tasks.

Recently, SimGRACE \cite{10.1145/3485447.3512156} performed encoder perturbation instead of augmenting or corrupting the graph view. The input graph feeds into a GNN encoder and its perturbed version. Then, the normalized cross entropy loss \cite{NEURIPS2020_3fe23034,\cmmnt{8578491}pmlr-v119-chen20j,ijcai2021p204} is applied for the two graph representations resulting from the GNN encoder and its perturbed version to maximize the agreement between positive pairs compared with negative pairs.

\section{Proposed Approach}
\label{sec:proposed}
In this section, we first outline the proposed endoscopic image matching approach before detailing our model architecture as well as the training process based on the cross-view contrastive learning.

\subsection{Image matching approach}
\begin{figure}[t]
\centering
\includegraphics[scale=0.8]{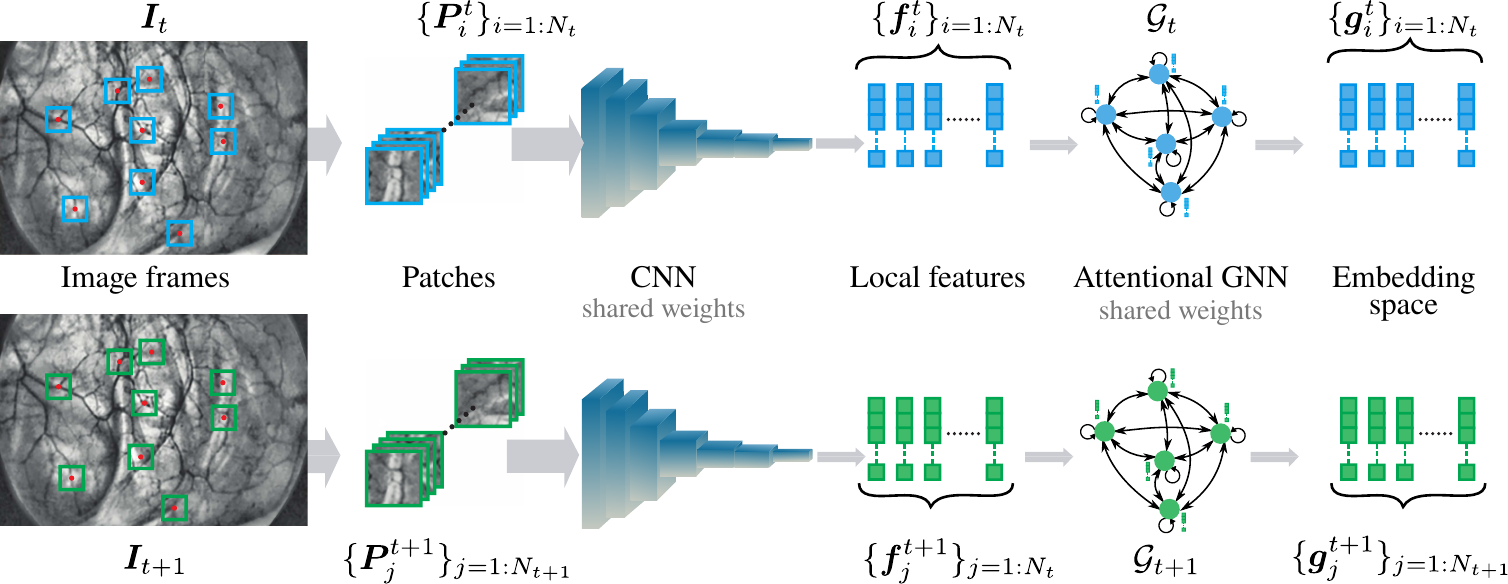}
\caption{The general principal of the proposed endoscopic image matching process.}
\label{Fig:1}
\end{figure}
The general principal of the proposed approach is depicted in \figureabvr \ref{Fig:1}. Given an input of two consecutive endoscopic images $\Ib_{t}$ and $\Ib_{t+1}$, the first step consists of applying a standard image interest point detector yielding to two sets of {2-D} points noted $\{\pb^{t}_{i}\}_{i=1:N_{t}}$ and $\{\pb^{t+1}_{j}\}_{j=1:N_{t+1}}$, where $N_{t}$ and $N_{t+1}$ are the cardinals of the two point sets detected in $\Ib_{t}$ and $\Ib_{t+1}$, respectively. The ultimate goal of the matching approach is to find corresponding points between these two sets. To that end, we propose extracting discriminative features that not only capture each point's local appearance but also provide a more global representation based on the point neighborhood. Indeed, we extract squared image patches $\{\Pb^{t}_{i}\}_{i=1:N_{t}}$ and $\{\Pb^{t+1}_{j}\}_{j=1:N_{t+1}}$ of $128\times128$ pixels centered on the detected points. These image patches are then fed into a CNN, which produces discriminative local feature vectors noted by $\{\fb^{t}_{i}\}_{i=1:N_{t}}$ and $\{\fb^{t+1}_{j}\}_{j=1:N_{t+1}}$. In order to leverage the spatial relationship between a given points and its neighboring points, we mapped all points to a fully connected graphical structures $\Gc_{t}$ and $\Gc_{t+1}$, each image separately. Graph nodes are represented by their point coordinates and feature vectors, for instance the $i$-th node in $\Gc_{t}$ is represented by $\pb^{t}_{i}$ and $\fb^{t}_{i}$.  An attentional Graph Neural Network is then used to transform $\Gc_{t}$ and $\Gc_{t+1}$ to produce the global feature vectors representing each reference point, resulting in $\{\gb^{t}_{i}\}_{i=1:N_{t}}$ and $\{\gb^{t+1}_{j}\}_{j=1:N_{t+1}}$ corresponding to $\{\pb^{t}_{i}\}_{i=1:N_{t}}$ and $\{\pb^{t+1}_{j}\}_{j=1:N_{t+1}}$, respectively.

The last step in the point matching approach is performed based on the Euclidean distance between points represented by their global features. That is, a given {$i$-th} point detected in $\Ib_{t}$ is matched to {$\hat{j}$-th} from $\Ib_{t+1}$ following equation \ref{eq:0}.
\begin{equation}
\hat{j}=\underset{j=1:N_{t+1}} {\mathrm{argmin}} \left\Vert \gb^{t}_{i} - \gb^{t+1}_{j}\right\Vert_2
\label{eq:0}
\end{equation}
\subsection{Model architecture}
\subsubsection{Local Feature Extraction}

Inspired by L2-Net \citep{8100132}, which was adopted widely by several works for local feature extraction \citep{sosnet2019cvpr,\cmmnt{8578167}Luo2018GeoDescLL}, our architecture is composed of seven convolutional layers. Each convolutional layer, except the last one, has a small kernel size $(3\times3)$ and is followed by batch normalization and ReLU. The last layer has a kernel size of $(8\times8)$ and it is followed by a batch normalization. 
\begin{table}[bh!]
\caption{Description of the proposed CNN architecture. Our architecture is composed of seven convolutional layers. The number of filters used for each  convolution layer is respectively  $\left\{16, 16, 32, 64, 128, 128, 128\right\}$. We applied padding with zeros on the first six layers to preserve the spatial size. We also used a convolution stride of 2 instead of using pooling layers to reduce the spatial size.}
\label{table:1}
\begin{tabular}{ccccccc}
\toprule
Layer & Type & Kernel size & Filters & Stride & Padding & Output layer    \\ 
\midrule
1     & \makecell{ Convolution \\ Bn+ReLU}  & $3\times3$  & 16      & 1      & 1       & $128\times128\times16$ \\
2     & \makecell{ Convolution \\ Bn+ReLU} & $3\times3$  & 16      & 2      & 1       & $64\times64\times16$   \\
3     & \makecell{ Convolution\\ Bn+ReLU} & $3\times3$  & 32      & 2      & 1       & $32\times32\times32$   \\
4     & \makecell{ Convolution \\ Bn+ReLU} & $3\times3$  & 64      & 2      & 1       & $16\times16\times64$   \\
5     & \makecell{ Convolution \\ Bn+ReLU} & $3\times3$  & 128     & 2      & 1       & $8\times8\times128$    \\
6     & \makecell{ Convolution\\ Bn+ReLU} & $3\times3$  & 128     & 1      & 1       & $8\times8\times128$    \\
7     & \makecell{ Convolution\\ Bn}      & $8\times8$  & 128     & 1      & -       & $1\times128$   \\ 
\bottomrule
\end{tabular}
\end{table}

Our CNN architecture requires a $(128\times128)$ patch as its input to provide a 128-D embedding vector as output. \tableabvr \ref{table:1} summarizes the details of the proposed architecture.

\subsubsection{Attentional Graph Neural Network}
For a sake of simplicity, we purposefully omit frame indices $t$ and $t+1$ from the following description as both $\Gc_{t}$ and $\Gc_{t+1}$ are processed in a similar way. The initial feature for each node is provided by a positional encoder which combines its local visual appearance $\fb_i$ extracted by the CNN and its location $\pb_i$ in the image coordinate system. The positional encoding allows the system to leverage both position and visual appearance, especially when combined  with an attention mechanism, such as the positional encoder used in transformers for language processing applications \cite{pmlr-v70-gehring17a,10.5555/3295222.3295349}.
Then, to propagate information along all edges, we compute the message passing based on an attention propagation block\citep{10.5555/3305381.3305512,sarlin2020superglue}. That is, the design of the Attentional Graph Neural Network is composed of two blocks: the positional encoder and the attentional aggregation.

\noindent\underline{\textit{\textbf{Positional encoding:}}} The positional encoding gives each node unique position information known as Node Positional Feature. It uses a Multilayer Perceptron (MLP) to embed the initial reference point {2-D} coordinates $\pb_{i}$ into a high dimensional vector. By adding this information to the visual feature $\fb_i$ provided by the CNN, the node feature becomes position-dependent. The initial representation of the $i$-th node noted by $^{0}\!\fb_i$ is obtained following equation \ref{eq:1}.
\begin{equation}
{}^{0}\!\fb_i=\fb_i+MLP(\pb_i)
\label{eq:1}
\end{equation}

\noindent\underline{\textit{\textbf{Attentional Aggregation:}}} Inspired by SL-ViT\cite{BAKHTIARNIA2022461}, we used a single attention layer. Thus, the update representation ${}^{1}\!\fb_i$ is computed as follow:
\begin{equation}
{}^{1}\!\fb_i = {}^{0}\!\fb_i + MLP\left(\left[{}^{0}\!\fb_i | \mb_i\right]\right)
\label{eq:2}
\end{equation}
\noindent where the $|$ operator applied a concatenation and $\mb_i$ is the message passing vector computed by the aggregation from all nodes to the $i$-th node.
This message passing process is computed based on a transformer encoder \cite{10.5555/3295222.3295349}, where the attention layer is a fundamental element. It takes input in the form of three vector parameters commonly known as \textit{Query} ($Q$), \textit{Key} ($K$), and \textit{Value} ($V$) \cite{10.5555/3295222.3295349}. The attention layers perform the aggregation and compute the message $\mb_i$ calculated by the aggregation from all nodes in the graph following equation \ref{eq:3}.
\begin{equation}
\begin{split}
\mb_i&=\sum_{l:1..N}Softmax(Q_{i}^{\mathsf{T}}K_{l})V_{l} \\
\text{where:}& \\
Q_{i} &= \Wb_{\Scale[0.5]{Q}} \;\; {}^{1}\!\fb_i + \bb_{\Scale[0.5]{Q}} \\
K_{l} &= \Wb_{\Scale[0.5]{K}} \;\; {}^{1}\!\fb_l + \bb_{\Scale[0.5]{K}}\\
V_{l} &= \Wb_{\Scale[0.5]{V}} \;\; {}^{1}\!\fb_l + \bb_{\Scale[0.5]{V}}
\end{split}
\label{eq:3}
\end{equation}
Computing $Q_{i}$, $K_{l}$, and $V_{l}$ in Equation \ref{eq:3} involves performing linear projections on node features. The $\Wb_{\Scale[0.5]{Q}}$, $\Wb_{\Scale[0.5]{K}}$, $\Wb_{\Scale[0.5]{V}}$, $\bb_{\Scale[0.5]{Q}}$, $\bb_{\Scale[0.5]{K}}$, and $\bb_{\Scale[0.5]{V}}$ variables stand in for the learnable weights and biases for these projections.

\subsection{Cross-View Contrastive Learning}
To train our model, we developed a novel self-supervision pipeline inspired by \cite{Zhu:2020vf} that employs Cross-View Contrastive Learning. The key concept of contrastive learning is to identify positive pairs and differentiate between negative pairs in the embedding space produced by the attention GNN in our case. To achieve this, our first objective is to generate appropriate positive and negative pairs. In supervised approaches that use annotated datasets, the positive pairs are selected from the same class as the anchor element, and the negative pairs are selected from other classes. For an optimal performance, hard negatives that are as close as possible to the anchor are typically preferred \cite{Schroff_2015}. In self-supervised contrastive learning, we use data augmentation to select positive pairs for the anchors or co-occurrence as described in \cite{pmlr-v119-chen20j,\cmmnt{hjelm2018learning}pmlr-v119-henaff20a}.

Contrastive learning has been widely used to learn graph representations in various studies \cite{DeepWalk,10.1145/2939672.2939754,10.1007/978-3-319-93417-4}.
In \cite{Zhu:2020vf}, the authors proposed a contrastive learning approach to minimize the agreement of node embedding between two graphs. They adopted a node training strategy based on two correlated graph views generated by intentional node feature corruption and/or edge removal.

\begin{figure}[h]
\centering
\includegraphics[scale=0.7]{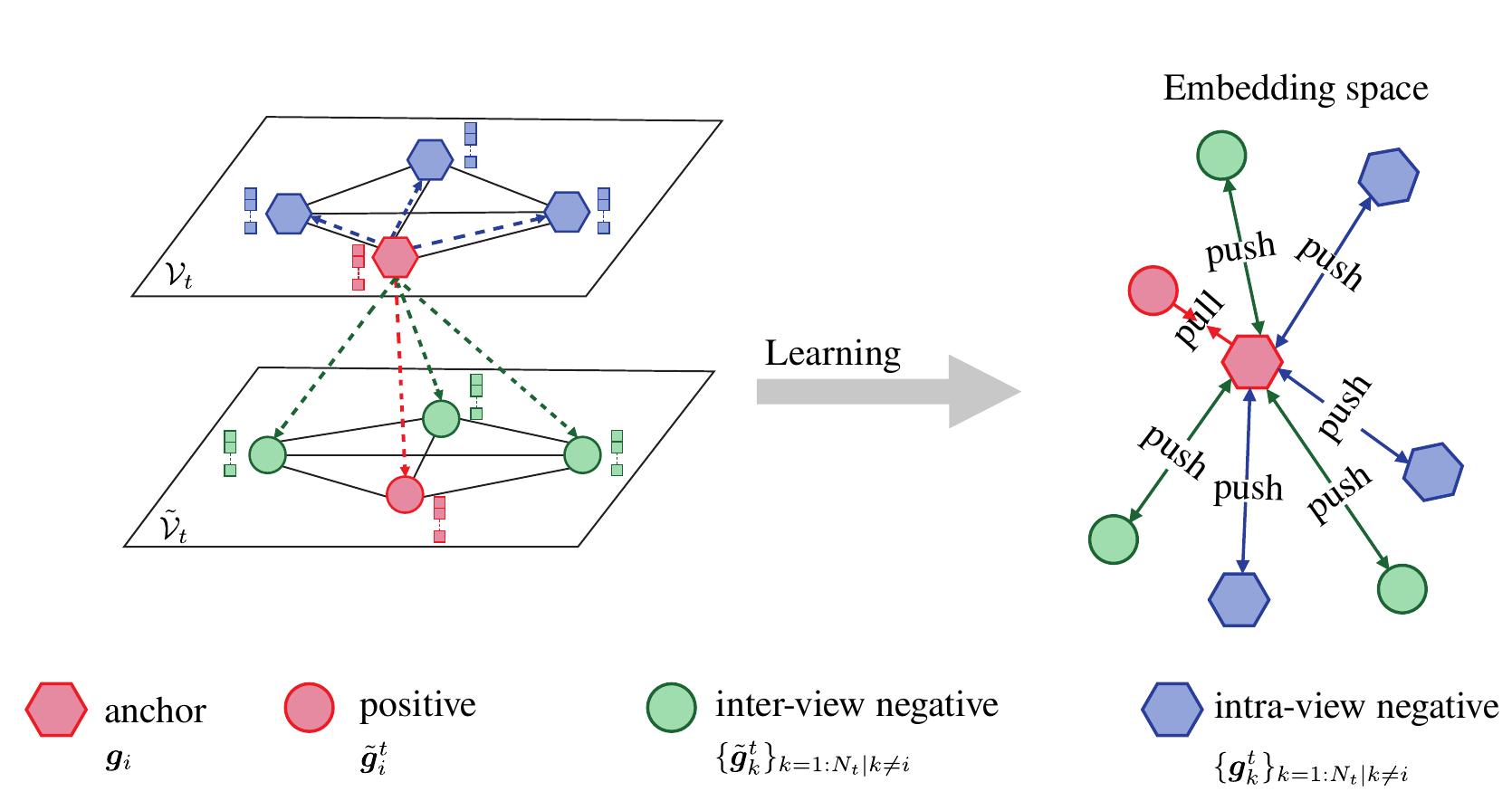}
\caption{Cross-view graph contrastive learning design. The red dashed line represents the distance in the embedding space between the same nodes in two augmented views (anchor and positive sample). The blue and the green dash lines represent the distance between the anchor and negatives samples which are provided respectively from two augmented views: inter-view or intra-view negatives.}
\label{Fig:2}
\end{figure}
We use arbitrary affine transformations $T(\cdot)$ to construct augmented graphs from an input image, which differs from the Grace framework \cite{Zhu:2020vf}. Specifically, we considered a given image $\Ib_t$ as the first view denoted by $\Vc_t$, and generated an augmented view $\Tilde{\Vc}_t$ by transforming $\Ib_t$ with $T(\cdot)$. The nodes and underlying feature vectors in $\Vc_t$ were computed using the attentional GNN. Similarly, the augmented view $\Tilde{\Vc}_t$ consisted of the same nodes as $\Vc_t$, but with their 2D positions altered by the affine transformation $T(\cdot)$, and their feature vectors updated accordingly since the input image had changed to $T(\Ib_t)$.

As illustrated in \figureabvr \ref{Fig:2}, an anchor could be any node $\gb_i$ from $\Vc_t$ and its corresponding positive node in $\Tilde{\Vc}_t$  is denoted by $\Tilde{\gb}^t_i$. Intra-view negative nodes are all other nodes $\{\gb^{t}_{k}\}_{k=1:N_{t} | k\neq i}$ in $\Vc_t$, while their corresponding nodes in $\Tilde{\Vc}t$ are called inter-view negative nodes, denoted by $\{\Tilde{\gb}^{t}_{k}\}_{k=1:N_{t} | k\neq i}$.

Based on these definitions, we formulated the contrastive loss as minimizing the distance between the anchor and its positive node in the embedding space, while simultaneously maximizing the distance between the anchor and all inter-view and intra-view negative nodes. Thus, the loss is defined as:

\begin{equation}
L(\gb^{t}_{i},\Tilde{\gb}^{t}_{i}) = -log \frac{e^{ \frac{\Theta(\gb^{t}_{i},\Tilde{\gb}^{t}_{i})}{\tau} }}{ \sum_{k=1_{[k\neq i]}}^{N_t}e^{  \frac{\Theta(\gb^{t}_{i},\gb^{t}_k)}{\tau} }+\sum_{k=1_{[k\neq i]}}^{N_t}e^{  \frac{\Theta(\gb^{t}_{i},\Tilde{\gb}^{t}_{k})}{\tau} }}
\label{eq:6}
\end{equation}

\noindent where $\tau$ is a normalization scale factor to be set experimentally and $\Theta(\gb^{t}_{i},\Tilde{\gb}^{t}_{i})$ is the cosine similarity between the non-linear projection of $\gb^{t}_{i}$ and $\Tilde{\gb}^{t}_{i}$ performed by a two MLP layers. This additional non-linear projection improves the expression power of the cosine similarity as claimed in \cite{Zhu:2020vf,pmlr-v119-chen20j}. The final loss $\Lc$ when applied to all nodes in the two views  $\Vc_t$ and  $\Tilde{\Vc}_t$ is computed as follows: 
\begin{equation}
\Lc=\frac{1}{2N_t}\sum_{i=1}^{N_t}( L(\gb^{t}_{i},\Tilde{\gb}^{t}_{i}) + L(\Tilde{\gb}^{t}_{i}, \gb^{t}_{i}) )
\end{equation}

\section{Experiments and results}
\label{sec:experiments}
We evaluate our approach for endoscopic image matching in regards of different aspects, including:

\begin{itemize}
    \item Intrinsic evaluation of our proposed method: we investigate the robustness of our descriptor against geometric and photometric transformations between frames.
    \item Comparison to state-of-the-art deep learning-based methods for local feature descriptor.
    \item Use-case of endoscopic image mosaicing: we evaluate qualitatively the efficiency of our method in the use case of endoscopic image mosaicing.   
\end{itemize}

Before delving into the details of these experiments, we first describe our dataset as well as the training settings.

\begin{figure}[h]
\centering
\includegraphics[width=0.8\textwidth]{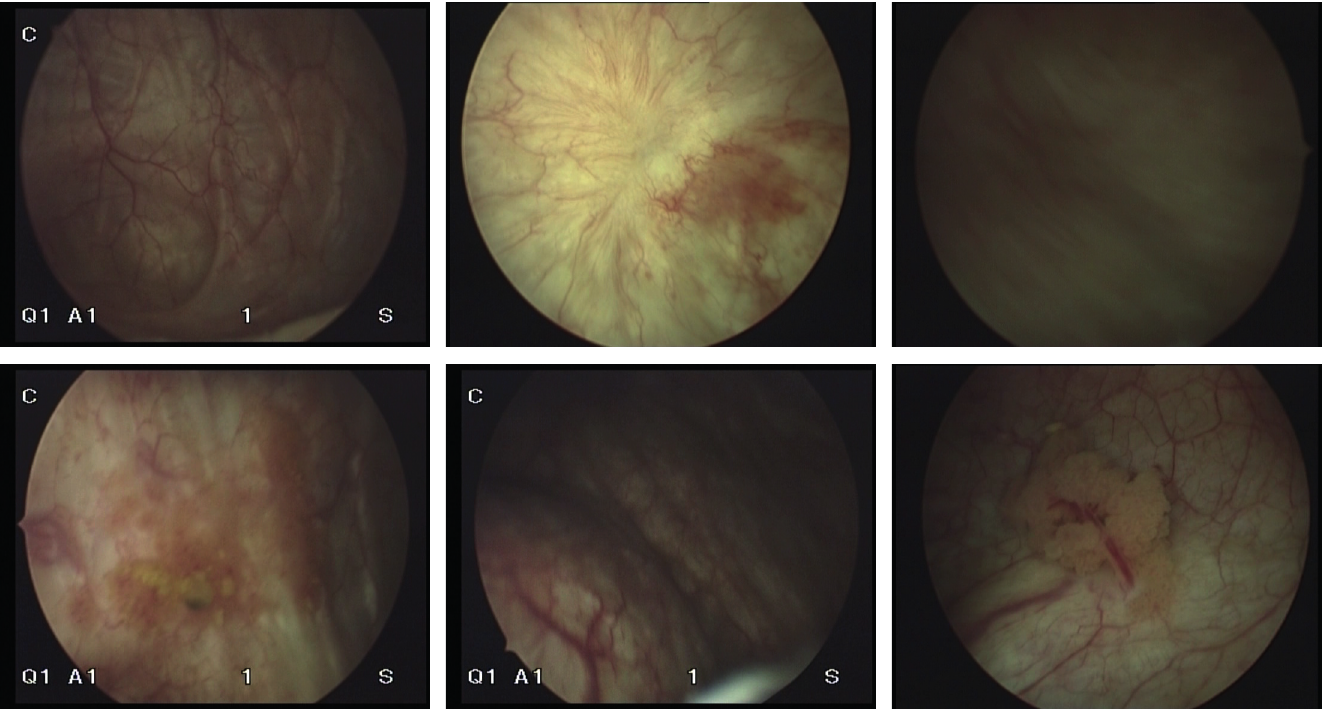}
\caption{Samples of clinical endoscopic images taken from different patients.}
\label{Fig:3}
\end{figure}

Full clinical endoscopic videos taken for 21 patients were used for our experiments. \figureabvr \ref{Fig:3} shows few frame samples from the dataset. In all of the conducted experiments, we randomly selected 16 video sequences to generate our training data and  the remaining 5 video sequences for validation in a cross-validation scheme. In addition, we repeated all experiments 5 times with different random splits in a cross-validation scheme.
\subsection{Database description}
Our training database consists of a collection of graph views derived from endoscopic video frames. To generate these graph views, we initially convert the endoscopic data into grayscale images and enhance the image contrast using the CLAHE method \cmmnt{\cite{6968381}}. Next, we employ a handcrafted method to detect key-points in each image, and we extract a $128\times128$ patch centered around each key-points. These patches are then mapped to an ergodic graphical structure, where each node represents a key-points characterized by its position and corresponding patch. To create two correlated graph views, we apply a transformation function $T(·)$ to the input image and randomly select one of the following transformations: small rotations with angles of $\theta=[5,10,15]$ degrees, small translations in both axes ($[4,6,8,10]$ pixels), and scaling with small scale factors ($S_{f}=[0.9, 0.95, 1.05, 1.1, 1.15]$). These transformations are inspired by the Grace framework \cite{Zhu:2020vf}.

\subsection{Training settings}
The database was trained using the Adam optimizer with a constant leaning rate of $5.10^{-4}$. Our model converges in about 48 hours of training on a NVIDIA Titan V GPU with 12GB memory. We fixed experimentally the normalization scaling factor $\tau$ of equation \ref{eq:6} to 0.08, and the mini-batch of node neighbors equal 10. The matching process between image pairs of size $576\times720$ is achieved in less than 1 second.

\subsection{Intrinsic evaluation of the proposed method}
In this section, we evaluate the performance of the proposed descriptor by conducting intrinsic evaluations against various geometric and photometric transformations. We utilize two matching methods, namely the Nearest Neighbor (NN) and the Nearest Neighbor with a Threshold (NNT) \citep{LeePark2015}, to assess the effectiveness of the descriptor. The NN method finds the closest match for each feature vector in the embedding space based on the Euclidean distance metric. On the other hand, the NNT method sets a predetermined threshold to accept matches with distances below this threshold. These matching methods are commonly employed to evaluate the performance of descriptors by measuring the similarity between patches using their feature vectors in the embedding space.

To access the matching performance, we used precision-recall and 1-precision curves. \citep{MikolajczykSchmid2005}. However, instead of computing the overlap error as in \citep{MikolajczykSchmid2005}, we used the projection error (PE) to determine correct and false matches. PE is calculated as the Euclidean distance between the detected key-points and the ground truth key-points, and we set the PE to 5 for all experiments.

We also compared our results to those obtained using well-known handcrafted descriptors, such as AKAZE, KAZE, SIFT, and ORB, each combined with its respective key-point detector. As each handcrafted descriptor is designed for a specific key-point detector, we report results for each of these detectors to ensure a fair comparison. For instance, when we use the provided code for AKAZE implementation to detect key-points, we refer to our descriptor as Proposed$_{AKAZE}$.

\subsubsection{Robustness against individual transformations}
\begin{figure}[t]
\centering
\includegraphics[width=0.99\textwidth]{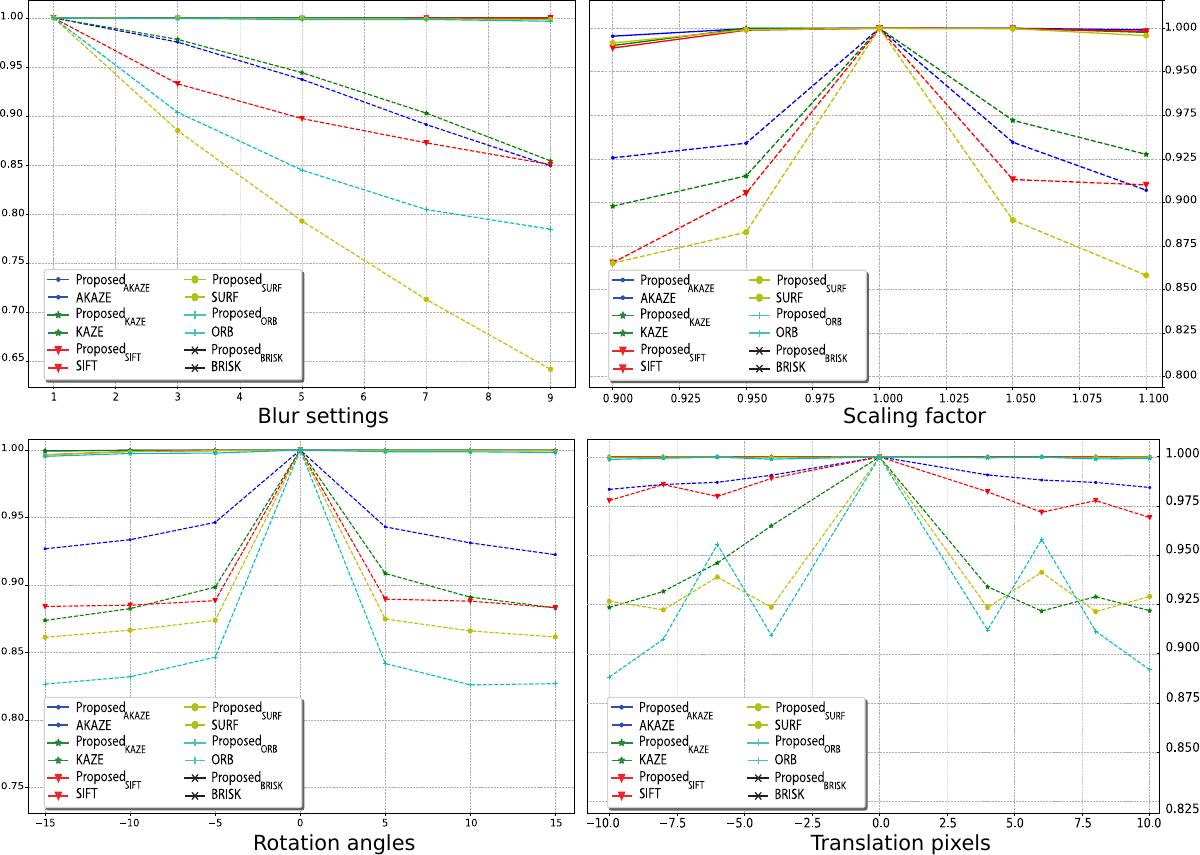}
\caption{Precision rates obtained for different individual transformations, including translation, rotation, scale change, and blur. The results demonstrate that the proposed descriptor achieves the highest precision rates, outperforming the performance of the handcrafted methods.}
\label{fig:4567}
\end{figure}

We evaluated the robustness of our proposed descriptor against four principal transformations: translation, rotation, scale change, and blur. To simulate these variations, we applied small rotations with angles ranging from 5 to 15 degrees ($\theta \in [5, 10, 15]$ degrees), small translations in both axes with pixel values in the range of 4 to 10 pixels, scaling with small scale factors ($S_{f}$=[0.9, 0.95, 1.05, 1.1, 1.15]), and motion blurring using kernel masks of different sizes ($3\times3$, $5\times5$, $10\times10$, and $15\times15$). Since the variations between consecutive frames in endoscopic videos are typically small, our focus was on assessing the performance under these small variations. The matching process was performed using the Nearest Neighbor (NN) approach. The precision curves obtained from these evaluations are summarized in Figures \ref{fig:4567}

Through the above quantitative analyses conducted above, the obtained results demonstrate the robustness of the proposed descriptor to individual transformations such as translation, rotation, scale change, and blur. It outperforms the state-of-the-art handcrafted methods in endoscopic image matching with the highest precision rates that are close to 1.

\subsubsection{Robustness against combined transformation}
In addition to assessing the robustness of the proposed descriptor against individual transformations, we also examined whether combined transformations would significantly impact its performance.

\begin{figure}[t]
\centering
\includegraphics[width=0.75\textwidth]{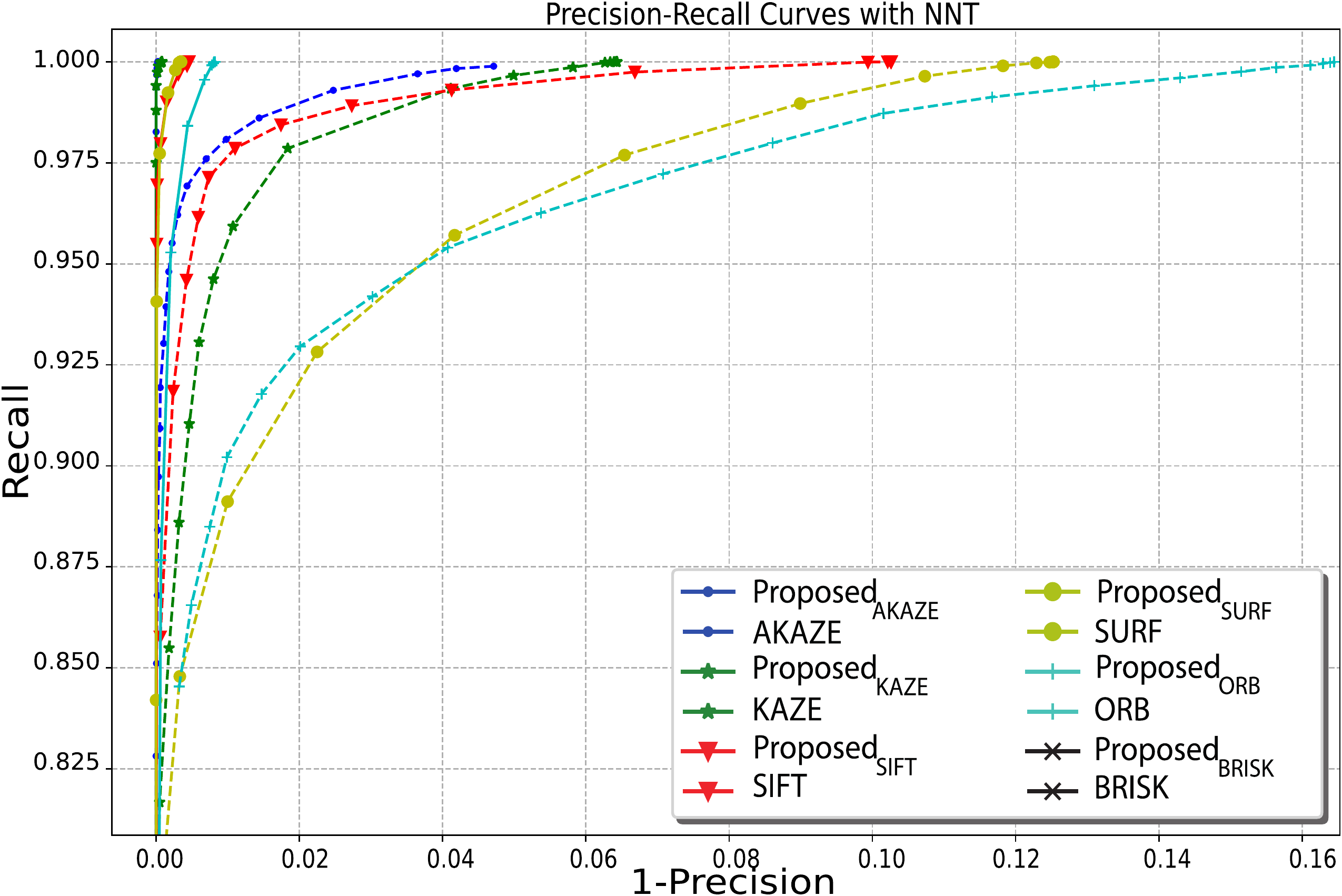}
\caption{Recall versus 1-precision curves for different viewpoints.}
\label{Fig:8}
\end{figure}

In the first test,  we focused on evaluating the method's ability to handle viewpoint changes, which involve a combination of geometric transformations. To simulate viewpoint changes in endoscopic videos, we estimated homographies between consecutive frames using the SIFT descriptor and the RANSAC algorithm. 
These estimated homographies served as ground truth for generating the validation dataset. For each frame in the validation endoscopic videos, we randomly applied a transformation chosen from the ground truth homographies. Then, for each pair of images (current image and warped image), we detected key-points, extracted local features, and matched descriptors. We used the nearest neighbor with a threshold (NNT) as the matching strategy. By varying the threshold and computing the recall and 1-precision, we obtained the curves shown in Figure \ref{Fig:8}, which summarize the performance of the proposed method under viewpoint changes.

We can observe that handcrafted descriptors are more sensitive to viewpoint change. Our descriptor outperforms handcrafted methods. The maximal rate of recall is reached for a precision rate of about 1 for Proposed$_{KAZE}$ and Proposed$_{AKAZE}$ setting. The obtained result shows also that the AKAZE descriptor outperforms the other handcrafted methods.

\subsection{Comparison to deep learning-based methods}

To validate the effectiveness of our descriptor, we compare its performance in terms of precision and matching score to the latest state-of-the-art deep learning methods for key-point image matching. Specifically, we compare our descriptor to the supervised-based methods SosNet \citep{sosnet2019cvpr}, SuperGlue \citep{sarlin2020superglue}, HyNet \citep{hynet2020}, and CNDesc \citep{CNDesc9761761}, as well as to the self-supervised methods proposed in our previous study \citep{FARHAT2023118696,chaabouni2022deep}. To ensure a fair comparison, we used SIFT as key-point detection baseline for all the compared methods.

The comparative results in terms of precision and matching score are presented in Table \ref{table:2}.

The obtained results clearly demonstrate the superiority of our descriptor, as it achieves the highest precision and matching score. Despite being trained in a self-supervised manner, our descriptor outperforms the four tested supervised methods, namely SosNet \citep{sosnet2019cvpr}, SuperGlue \citep{sarlin2020superglue}, HyNet \citep{hynet2020}, and CNDesc \citep{CNDesc9761761}. Additionally, it outperforms our own self-supervised descriptor proposed in the previous study \citep{FARHAT2023118696}, highlighting the significance of incorporating spatial and topological relationships for achieving high precision. These results underscore the importance of leveraging spatial relationship knowledge to enhance the performance of our descriptor.

\begin{table}[h!]
\caption{comparative results in terms of precision and matching score between our method and the most recent state-of-the-art deep learning methods for key-point image matching.}
\label{table:2}
\centering
\begin{tabular}{lcc}
\hline
& \textit{Precision}     & \textit{Matching Score}  \\ \hline
SosNet \citep{sosnet2019cvpr}& 0.9995            & 0.9160  \\
SuperGlue \citep{sarlin2020superglue}    & 0.9917            & 0.9281 \\
HyNet \citep{hynet2020}   & 0.9996           & 0.9029            \\
CNDesc \citep{CNDesc9761761}  & 0.9837            & 0.9051           \\
 \citep{FARHAT2023118696}   & 0.9989   & 0.9256 \\
 Proposed & \textbf{1}  & \textbf{0.9930} \\
 \hline
\end{tabular}
\end{table}

\subsection{Use-case of endoscopic image mosaicing}
\begin{figure}[h!]
\centering
\includegraphics[width=1\textwidth]{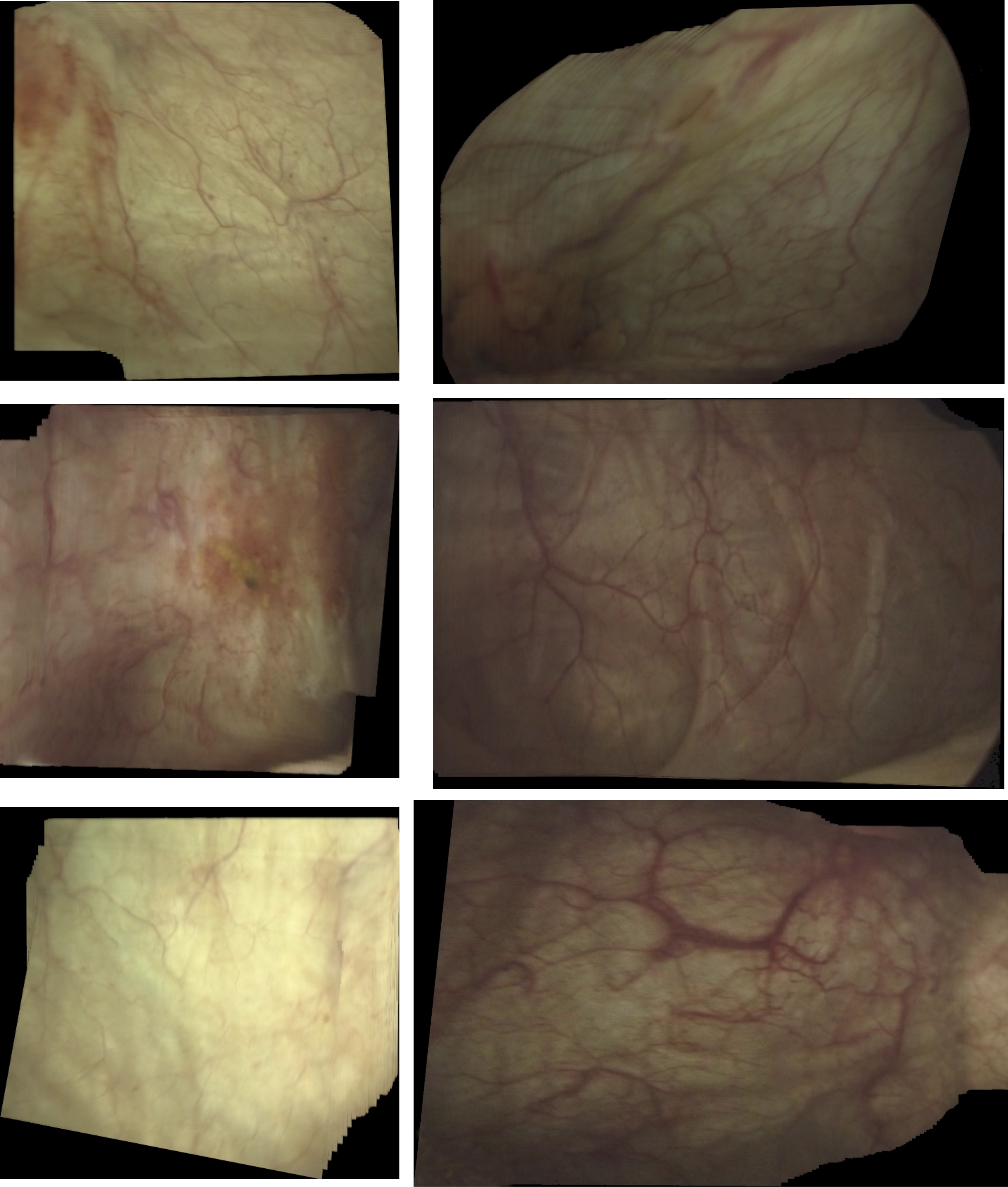}
\caption{Examples of extended mosaics constructed for different endoscopic sequences or different lengths.}
\label{Fig:9}
\end{figure}

To analyze the performance of our proposed descriptor in generating panoramic images, we conducted additional experiments. We utilized the matching results obtained from our descriptor and applied them in a image mosaicing baseline based on the well-known RANSAC algorithm. 
We generated a panoramic image for different sequences with different lengths ranging between 200 to 800 images. The resulting panoramic images are shown in Figure \ref{Fig:9}. The resulting extended mosaic images exhibited seamless mosaicing with no visible texture discontinuities. The continuity of the vessel patterns is clearly observable, indicating accurate alignment of the images. These samples highlight the effectiveness of our proposed descriptor in generating visually coherent and accurate panoramic images endoscopic videos. However, it is important to note that significant scale changes or strong blur effects can introduce disruptions and inconsistencies in the mosaicing process, potentially leading to visible discontinuities in the panoramic image. To mitigate this issue, an artifact detection algorithm can be integrated into the mosaic image system to identify and address any artifacts or inconsistencies that may arise.







\subsection{Further Analysis}
In contrastive loss learning, the temperature parameter plays a crucial role in determining the trade-off between tolerance and uniformity. 
To examine the impact of this factor on our model's performance in a matching task, we experiment with different values of the temperature parameter ($\tau$) previously introduced in Equation \ref{eq:6} using evenly spaced intervals. The precision and matching score rates obtained for each value of $\tau$ are summarized in \tableabvr \ref{table:3}.

\begin{table}[h!]
\caption{Matching performance at different values of the contrastive loss temperature parameter $\tau$.}
\centering
\label{table:3}
\begin{tabular}{lcccc}
\hline
         $\tau$    & 0.06 & 0.08 & 0.1 & 0.12 \\ \hline
\multicolumn{1}{l}{Precision}      &       0.9989      & 1           & 0.9999      &          0.9999   \\ \hline
\multicolumn{1}{l}{Matching Score} &        0.9841     & 0.9930       & 0.9943      &    0.9881         \\ \hline
\end{tabular}
\end{table}

The obtained results indicate that both small and large values of the temperature parameter $\tau$ result in inferior matching performance. Among the tested values, the model achieved the best performance with a temperature parameter of $\tau = 0.08$. Additionally, we examined the impact of another significant hyper-parameter: the mini-batch size. This parameter determines the number of inter-view and intra-view negative samples. By varying the mini-batch size at equal intervals, we assessed its effect on the model's performance. As shown in Table \ref{table:4}, our model consistently achieved high precision rates across a range of mini-batch sizes. However, it should be noted that smaller mini-batch sizes tended to yield slightly inferior precision, although the effect was not significant. Conversely, as the mini-batch size increased, the matching score decreased. This indicates that increasing the number of negative samples in graph contrastive loss learning does not necessarily lead to improved performance and may even result in training difficulties.

\begin{table}[h]
\caption{Matching performance under different values of mini-batch size hyper-parameter.}
\centering
\label{table:4}
\begin{tabular}{lcccc}
\hline
        Mini-batch size    & 5 & 10 & 15 & 20 \\ \hline
\multicolumn{1}{l}{Precision}      &      0.9998       & 1           & 1      & 1            \\ \hline
\multicolumn{1}{l}{Matching Score} &  0.9939           & 0.9930       & 0.9917      &   0.9817          \\ \hline
\end{tabular}
\end{table}
\section{Conclusions and perspectives}
\label{sec:conclusion}
In conclusion, we have presented a novel framework for endoscopic image key-point matching, which effectively combines CNN and GNN models to capture both visual appearance and spatial relationships. To address the scarcity of labeled data, we developed a graph self-supervised framework based on cross-view contrastive learning. Through extensive experiments, we have demonstrated the superiority of our method over existing supervised leaning methods, and the generated panoramic images showcased the precision of our descriptor in aligning images. However, there are still areas for further improvement, such as handling scale changes and blur effects in the mosaicing process and incorporating artifact detection algorithms. Additionally, conducting evaluations on larger datasets and exploring different GNN architectures and training strategies could provide valuable insights.

\bibliography{main}
\end{document}